\documentclass{article}
\usepackage{ijcai21}

\usepackage{times}
\usepackage{soul}
\usepackage{url}
\usepackage[hidelinks]{hyperref}
\usepackage[utf8]{inputenc}
\usepackage[small]{caption}
\usepackage{graphicx}
\usepackage{amsmath}
\usepackage{amsthm}
\usepackage{booktabs}
\usepackage{algorithm}
\usepackage{algorithmic}
\usepackage{comment}
\usepackage{enumitem}
\usepackage{subfigure}
\usepackage{multirow}
\usepackage{multicol}
\usepackage{bigstrut}
\usepackage{amssymb}
\usepackage{xcolor}

\newcommand{\etal}{\textit{et al}.~}

\newcommand{\ieno}{\textit{i}.\textit{e}.}

\newcommand{\egno}{\textit{e}.\textit{g}.} 

\newcommand{\etcno}{\textit{etc}} 

\title{Uncertainty-Aware Few-Shot Image Classification}

\author{
Zhizheng Zhang$^1$\footnote{This work was done when Zhizheng Zhang was a visiting student at Columbia University.}\and
Cuiling Lan$^2$\footnote{Corresponding Author}\and
Wenjun Zeng$^2$\and
Zhibo Chen$^{1\dagger}$\and
Shih-Fu Chang$^3$\\
\affiliations
$^1$University of Science and Technology of China\\
$^2$Microsoft Research Asia\\
$^3$Columbia University\\
\emails
zhizheng@mail.ustc.edu.cn\and
\{culan, wezeng\}@microsoft.com\\
chenzhibo@ustc.edu.cn\and
sc250@columbia.edu
}

\begin{document}

\maketitle

\begin{abstract}

Few-shot image classification learns to recognize new categories from limited labelled data. Metric learning based approaches have been widely investigated, where a query sample is classified by finding the nearest prototype from the support set based on their feature similarities. A neural network has different uncertainties on its calculated similarities of different pairs. Understanding and modeling the uncertainty on the similarity could promote the exploitation of limited samples in few-shot optimization. In this work, we propose Uncertainty-Aware Few-Shot framework for image classification by modeling uncertainty of the similarities of query-support pairs and performing uncertainty-aware optimization. Particularly, we exploit such uncertainty by converting observed similarities to probabilistic representations and incorporate them to the loss for more effective optimization. In order to jointly consider the similarities between a query and the prototypes in a support set, a graph-based model is utilized to estimate the uncertainty of the pairs. Extensive experiments show our proposed method brings significant improvements on top of a strong baseline and achieves the state-of-the-art performance.

\end{abstract}

\section{Introduction}

The strong capability of deep learning in part relies on the using of a large amount of labeled data for training, while humans are more readily able to learn knowledge quickly given a few examples. Few-shot learning \cite{vinyals2016matching,finn2017model,karlinsky2019repmet,dong2018few,mishra2018generative} strives to a further step to develop deep learning approaches which generalize well to unseen tasks/classes with just few labeled samples. Among them, few-shot image classification \cite{vinyals2016matching,snell2017prototypical,sung2018learning,oreshkin2018tadam,finn2017model,nichol2018first,lee2019meta} has attracted much attention, which aims to classify data (query samples) of new categories given only a few labeled samples of these categories as examples (support samples).

For few-shot image classification, several latest research works reveal that good feature embedding is important to deliver favorable performance for the similarity-based classification  \cite{chen2019closer,chen2020new,huang2019all,tian2020rethinking}. They average the feature embeddings of the support samples of a category to be the prototype of this category. Thus, with each prototype being category-specific, the set of prototypes plays the role of similarity-based classifier to identify the category of a query sample by finding the nearest prototype over the support set in the embedding space. During training, for a query sample, the logit vector (classification probability) is obtained by feeding the similarities between the query sample and the prototypes to a SoftMax function.
Hence, the reliability of the estimated similarity is vital to the classification performance.

The quality of network output has been investigated and modeled with uncertainty in regression, classification, segmentation, multi-task learning \cite{kendall2015bayesian,kendall2017uncertainties,chang2020data,kendall2018multi} to benefit the optimization. Aleatoric uncertainty identifies the confidence level of the network on its output for a sample, which captures the noise inherent in the observations. For few-shot image classification, the network actually has different uncertainties on the calculated similarities of different query-prototype pairs. An observed similarity of a query-prototype pair, being a one time sampling, suffers from observation noise, where the higher of the uncertainty, the less reliable of the estimated similarity. For each few-shot task during the optimization, the number of experienced query-prototype pairs is limited and thus the side effect of observation noises due to the high uncertainty limits the optimization. Therefore, modeling uncertainty is especially vital for few-shot learning with the limited samples, but still under-explored.

In this paper, we propose an efficient uncertainty-aware few-shot image classification framework to model uncertainty and perform uncertainty-aware optimization. For each query-prototype pair, we convert the observed feature similarity between the query and the prototype from a deterministic scalar to a distribution, with the variance of this distribution characterizing the uncertainty of the observed similarity. The similarity-based classification losses are calculated based on the Monte Carlo sampling on the similarity distributions to alleviate the influence of observation noises. Since the classification probability of a query sample is determined based on the similarity of this query sample with its $N$ prototypes in the support set, this is a joint determination process. Thus, we adopt a graph-based model to jointly estimate the uncertainties for the $N$ query-prototype pairs, which facilitates the information passing among them for the joint optimization.

In summary, our contributions lie in the following aspects:

\begin{itemize}[noitemsep,nolistsep,leftmargin=*]
\item We are the first to explore and model the uncertainty of the similarities of  query-prototype pairs in few-shot image classification, where a better understanding of the limited data is particularly important.

\item We perform uncertainty-aware optimization to make the few-shot learning more robust to observation noises.

\item We design a model to jointly estimate the uncertainty of the similarities between a query image and the prototypes in the support set.

\end{itemize}

We conduct comprehensive experiments and demonstrate the effectiveness of our proposed method in the popular inductive setting, where each query sample is evaluated independently from other query samples. 
Our scheme achieves the state-of-the-art performance on multiple datasets.

\section{Related Work}

\subsection{Few-shot Image Classification}
Few-shot image classification aims to recognize novel (unseen) classes upon limited labeled examples. Representative approaches can be summarized into four categories.

\noindent\textbf{Classification-based methods} train both a feature extractor and classifiers with meta-learning and learn a new classifier (\egno, linear softmax classifier) for the novel classes during meta-testing  \cite{chen2019closer,ren2019incremental,lee2019meta}. They in general need to update/fine-tune the network given a few-shot testing task.

\noindent\textbf{Optimization-based methods} exploit more efficient meta-learning optimization strategies for few-shot learning \cite{finn2017model,grant2018recasting,nichol2018first,lee2019meta}. MAML \cite{finn2017model} is a representative work, which trains models by aligning the gradients of several learning tasks, such that only a few iterations are needed to achieve rapid adaptation for a new task.

\noindent\textbf{Hallucination-based methods} enhance the performance by augmenting data, which alleviates data deficiency by using generated data. The generators transfer appearance variations \cite{hariharan2017low,gao2018low} or styles \cite{antoniou2017data} from base classes to novel classes. Wang \etal create an augmented training set through a generator which is trained end-to-end along with the classification~\cite{wang2018low}.

\noindent\textbf{Similarity-based/Metric-based methods} classify an unseen instance into its nearest class based on the similarities with a few labeled examples. They learn to project different instances to a metric space wherein the similarities among instances of the same category are larger than that of instances of different categories. Matching networks \cite{vinyals2016matching} propose to assign an unlabeled data with the label of its nearest labeled data in one-shot learning. Furthermore, the prototypical network \cite{snell2017prototypical} averages the features of several samples of the same category as the prototype of this class and classify unseen data by finding the nearest prototype in the support set, where a prototype is taken as the representation of a class. Relation networks \cite{sung2018learning} propose an additional relation module which learns to determine whether a query and a prototype belongs to the same class and is jointly trained with deep feature representations.

Some recent metric-based works reveal the importance of learning good feature representations via classification-based pre-training for few-shot tasks \cite{chen2019closer,tian2020rethinking,chen2020new}. They propose strong baselines by training the network with classification supervision over all base categories and further performing meta-learning on sampled categories to simulate the few-shot setting. Chen \etal confirmed that the classification-based pre-training can provide extra transferability from base classes to novel classes in the meta-learning stage \cite{chen2020new}.

In this paper, on top of such a strong baseline, we look into an important but still under-explored issue, \ieno, modeling and exploiting the inherent uncertainty of the calculated similarities for more effective optimization.

\subsection{Uncertainty in Deep Learning}
There are two main types of uncertainty studied for deep neural networks: aleatoric uncertainty, and epistemic uncertainty \cite{kendall2017uncertainties,gal2016uncertainty,kendall2018multi}. Epistemic uncertainty captures model-related stochasticity, including parameters choices, architecture, \etcno. Aleatoric (Data-dependent) uncertainty captures the noise inherent in the observations \cite{kendall2017uncertainties,gal2016uncertainty,kendall2018multi}, which depends on the input sample of the model. A model usually has high uncertainty on noisy input or rarely seen input.

\begin{figure*}[t]
	\begin{center}
		\includegraphics[width=1\linewidth]{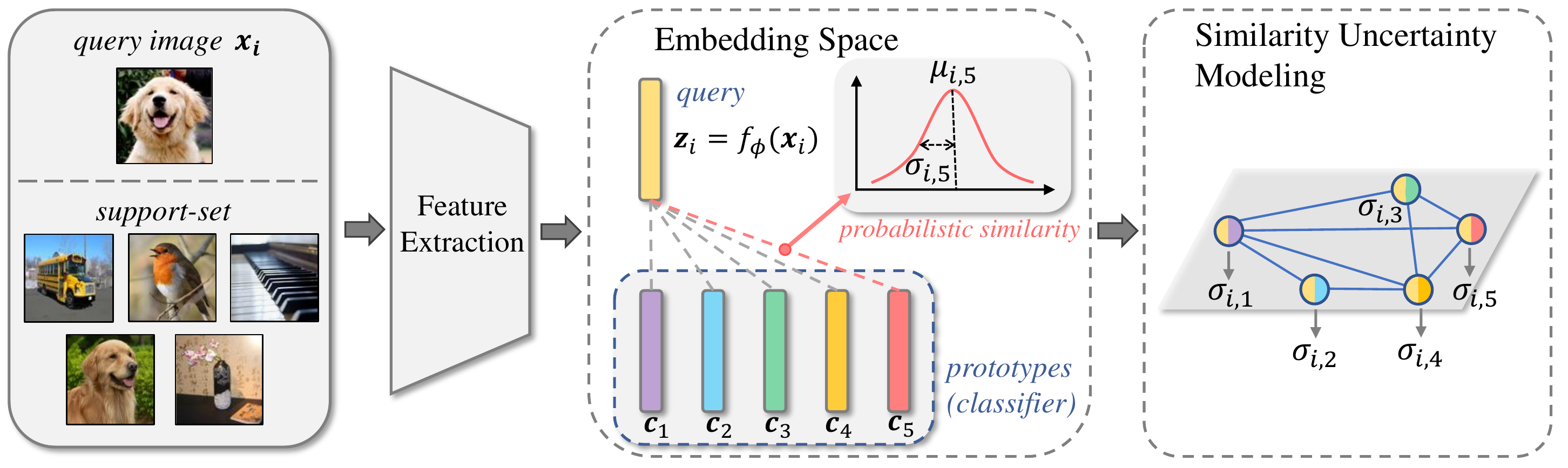}
	\end{center}
	\vspace{-2mm}
	\caption{\textbf{Pipeline of proposed Uncertainty-Aware Few-Shot (UAFS) image classification.} To alleviate the influence of observation noises on the similarity of a query-prototype pair, rather than a scalar, we model the similarity between a query feature $\mathbf{z}_i$ and a prototype feature $\mathbf{c}_j$ (where $j=1,\cdots,N$) by a distribution, \ieno, \textit{probabilistic similarity}, based on graph-based similarity uncertainty estimation. For a query sample $\mathbf{x}_i$ and $N$ prototypes in the support set, we take each query-prototype pair as a node and employ a graph-based model to jointly infer the similarity uncertainty $\sigma_{i,j}$ for each query-prototype pair. The network is optimized with the similarity-based classification losses which exploit the estimated uncertainty (not shown in this figure, please see the subsection around Eq.~(\ref{eq:5})).}
	\label{fig:pipeline}
	\vspace{-3mm}
\end{figure*}

For few-shot classification, the ``hardness'' of a few-shot episode is quantified with a metric in \cite{dhillon2019baseline} but is only used to evaluate few-shot algorithms rather than improve the optimization. In \cite{wang2020instance}, a statistical approach is adopted to measure the credibility of predicted pseudo-labels for unlabeled samples, where the most trustworthy pseudo-labeled instances are selected to update the classifier in the semi-supervised or transductive few shot learning setting. There is still a lack of exploration of the uncertainty for more general inductive few-shot setting, where the network does not use/need testing samples for updating. In this paper, we model the data-dependent uncertainty of the calculated similarities between a query sample and the prototypes, and leverage it to better optimize the network. To our best knowledge, we are the first to explore the uncertainty of pair-similarity in few-shot classification towards more efficient optimization.

\section{Proposed Method}
We propose a Uncertainty-Aware Few-Shot (UAFS) image classification framework (see  Figure \ref{fig:pipeline}), which models uncertainty of the query-prototype similarity and perform uncertainty-aware optimization. We first present the problem formulation and a strong baseline in Section 3.1. Then, we elaborate our proposed uncertainty-aware few-shot image classification in Section 3.2.

\subsection{Preliminaries and a Strong Baseline}
\label{subsec:preliminary}

\paragraph{Problem Formulation.} For few-shot image classification, all the categories/classes of the dataset are divided into base classes $\mathcal{C}_{base}$ for training and novel classes $\mathcal{C}_{novel}$ for testing without class overlapping \cite{snell2017prototypical,chen2019closer,dhillon2019baseline}. In few-shot learning, episodic training is widely used. Each $N$-way $K$-shot task randomly sampled from base classes is defined as an episode, where the support set $\mathcal{S}$ includes $N$ classes with $K$ samples per class, and the query set $\mathcal{Q}$ contains the same $N$ classes with $M$ samples per class. Our method is applied to the metric-based few-shot learning which performs similarity-based classification based on matching the prototypes in the support set for a given query sample.

Given an image $\mathbf{x}_i$, we feed it to a Convolution Neural Network (CNN) to obtain a feature map, and perform global averaging pooling to get a $D$-dimensional feature vector $\mathbf{z}_i=f_{\phi}(\mathbf{x}_i) \in \mathbb{R}^D$ as the embedding result of $\mathbf{x}_i$, where $f_{\phi}(\cdot)$ denotes the feature extractor. The prototype of a class indexed by $k$ is calculated by averaging the feature embeddings of all the samples of this class in the support set as:
\begin{equation}\label{eq:1}
\mathbf{c}_k=\frac{1}{|S_k|}\sum_{(\mathbf{x}_i, \mathbf{y}_i)\in S_k}f_\phi(\mathbf{x}_i),
\end{equation}
where $\mathbf{x}_i$ denotes the sample indexed by $i$ and $\mathbf{y}_i$ denotes its label, $S_k$ denotes the set of samples of the $k$-th class in the support set $\mathcal{S} = \{S_1, \cdots, S_N\}$, $|S_k|$ denotes the number of samples in $S_k$. For a query image $\mathbf{x}$, the probability of classifying it into the class $k$ is:
\begin{equation}\label{eq:2}
p(\mathbf{y}_i=k | \mathbf{x}_i)=\frac{exp(\tau \cdot s(f_\phi(\mathbf{x}_i), \mathbf{c}_k))}{\sum_{j=1}^{N}exp(\tau \cdot s(f_\phi(\mathbf{x}_i), \mathbf{c}_j))},
\end{equation}
where the temperature parameter $\tau$ is a hyper-parameter, and $N$ is the number of classes in the support set (\ieno, $N$-way). The $s(f_\phi(\mathbf{x}_i), \mathbf{c}_k)$ represents the similarity between the given query-sample $\mathbf{x}_i$ and the prototype $\mathbf{c}_k$ of the class $k$. Here, the $N$ prototypes of $N$ classes in the support set construct a similarity-based classifier.

\paragraph{Baseline.} Many works~\cite{vinyals2016matching,snell2017prototypical,chen2019closer,li2019finding,chen2020new} reveal that the core of metric-based few-shot learning is to learn a good feature embedding function $f_\phi(\cdot)$ from base classes $\mathcal{C}_{base}$ which could generalize well on unseen classes. We follow the latest work \cite{chen2020new} to build a strong baseline with two-stage training:

Stage-1: \textit{Classification-based pre-training}. We employ a Fully-Connected (FC) layer as the classifier with cosine similarity over all base classes to perform pre-training. We train the feature extractor $f_\phi(\cdot)$ (backbone) and the classifier in an end-to-end manner using a standard cross-entropy loss.

Stage-2: \textit{Meta-learning}. For each episode/task, we sample $N$-way $K$-shot from base classes for training, which can be understood as a simulation of few-shot test process. In each task, $N \times K$ images and $N \times M$ images are sampled to constitute the support set and the query set, respectively. We adopt the cross-entropy loss formulated by $\mathcal{L} = -\sum_{i=1}^{N \times M}\mathbf{y}_i\cdot {\mathrm{log}}(p(\mathbf{y}_i|\mathbf{x}_i))$, where $p(\mathbf{y}_i|\mathbf{x}_i)$ is calculated in Eq.~(\ref{eq:2}) with $s(\cdot, \cdot)$ instantiated by cosine similarity.

\subsection{Uncertainty-aware Few-shot Learning}
A neural network has different uncertainties regarding its outputs for different input data and there will be observation noises for an instantiated output \cite{kendall2017uncertainties}. For few-shot image classification, the model also has uncertainty on the estimated similarities for the query-prototypes pairs. The small number of samples in each few-shot episode exacerbate the side effect of observation noises from uncertainty. To address this, we model the data-dependent uncertainty of similarities and leverage it for better optimizing the network.

\paragraph{Probabilistic Similarity Representation.}
To characterize the uncertainty of similarity of a query-prototype pair, we convert the similarity representation of this query-prototype pair from a deterministic scalar to a probabilistic representation, \ieno, a parameterized distribution.

Given a query sample $\mathbf{x}_i$ (with its feature $\mathbf{z}_i=f_\phi(\mathbf{x}_i)$ in the latent space) and a prototype $\mathbf{c}_j$ obtained by Eq.~(\ref{eq:1}), instead of using a deterministic scalar, we model the similarity as a probability distribution. Similar to other works \cite{kendall2017uncertainties,gal2016uncertainty}, we simply model the distribution of the similarity between this query and the prototype $\mathbf{c}_j$ by a Gaussian distribution with mean $\mathbf{\mu}_{ij}$ and variance $\mathbf{\sigma}_{ij}^2$:
\begin{equation}\label{eq:3}
p(s_{ij}|\mathbf{z}_i,\mathbf{c}_j)=\mathcal{N}(s_{ij};\mu_{ij}, \sigma_{ij}^{2}I),
\end{equation}
where $\mathcal{C} = \{\mathbf{c}_1,\cdots, \mathbf{c}_N\}$ denotes the $N$ prototypes in the support set, $\mathbf{c}_j$ is the j-th prototype within it.

We estimate the mean of the similarity by their inner product, \ieno, $\mathbf{\mu}_{ij}=\langle\mathbf{z}_i, \mathbf{c}_j\rangle$, which denotes the most likely value of the similarity between $\mathbf{z}_i$ and $\mathbf{c}_i$. The variance $\mathbf{\sigma}_{ij}^2$ is used to characterize the uncertainty of this pairwise similarity. We estimate it by a graph neural network, which will be elaborated in the \emph{subsection ``Similarity Uncertainty Estimation''}.

\paragraph{Uncertainty-aware Optimization.}
With the similarity of a query-support pair being represented by a distribution rather than a scalar, we incorporate such probabilistic similarity representation in the classification loss for better optimization, which could be more robust to the observation noises.

The analytic solution of integrating over these distributions for the optimization of classification loss function is difficult. Following \cite{kendall2017uncertainties}, we thus approximate the optimization objective by Monte Carlo integration. Particularly, Monte Carlo sampling is performed on the $N$ similarity distributions. For a given query image $\mathbf{x}_i$ and $N$ prototypes $\mathbf{c}_j, j=1,\cdots,N$ which form $N$ similarity pairs, we repeat $T$ random sampling over the similarity distributions for each query-prototype pair $s_{ij}$ (see (\ref{eq:4}) with $t=1,\cdots,T$) to obtain statistical results. We enable a differentiable representation of the probabilistic similarity for a query and prototype pair $s_{ij}$ by re-parameterizing it as:
\begin{equation}\label{eq:4}
s_{ij, t}=\mathbf{\mu}_{ij} + \mathbf{\sigma}_{ij}\epsilon_t,\quad\epsilon_t \in \mathcal{N}(0, 1).
\end{equation} 

For a given query image $\mathbf{x}_i$ with groundtruth label $\mathbf{y}_i=k$, we obtain its corresponding classification loss as:
\begin{equation}\label{eq:5}
\mathcal{L}{(\mathbf{x}_i, \mathbf{y}_i=k)}=-{\mathrm{log}}(\frac{1}{T}\sum_{t=1}^{T}(e^{s_{ik,t}}/\sum_{j=1}^{N}e^{s_{ij,t}})).
\end{equation}

\paragraph{Similarity Uncertainty Estimation.} We characterize the uncertainty by variance $\sigma_{ij}^2$ as mentioned before. Hereafter, we elaborate on how to estimate the uncertainty.

For a $N$-way similarity-based classification, the given query is compared with all the $N$ prototypes and the class of the prototype with the highest similarity value is taken as the predicted class, which is \emph{a joint determination process}.

The joint determination nature inspires us to estimate uncertainty parameter $\sigma_{ij}$ for each pair after taking a global glance of the $N$ pairs with $j=1,\cdots,N$. Therefore, we design a graph-based model as the similarity uncertainty estimator to jointly infer all the $\sigma_{ij}$ for a given query $\mathbf{x}_i$ and $N$ prototypes $\mathbf{c}_j$. This enables the exploration of global scope information for the current task to jointly estimate the similarity uncertainties. Besides, the graph-based design has another advantage, \ieno, its scalability to the number of categories. This allows us to pre-train the parameters of the graph-based model it in Stage-1. We will demonstrate its effectiveness by comparing it with other alternative designs: 1) using Convolutional Neural Network (CNN) to infer $\sigma_{ij}$ independently for each pair; 2) adopting Fully-Connected (FC) based model that is able to exploit global information for $N$ pairs but cannot handle the consistency classes in different episodics well.

Given a query and $N$ prototypes as illustrated in Fig.~\ref{fig:pipeline}, we build a graph of $N$ nodes, with the $j^{th}$ node denoted by the information of the query and the $j^{th}$ prototype. The output of the node is the predicted uncertainty of the similarity of this query-prototype pair. We feed a query image $\mathbf{x}_i$ to the feature extractor to get its embedding vector $\mathbf{z}_i =f_\phi(\mathbf{x}_i) \in \mathbb{R}^D$. Similarly, we get the $N$ prototypes of the support set as $\mathbf{c}_j \in \mathbb{R}^D, j\!=\!1,\!\cdots\!,N$. We use the group-wise relations/similarities between the query and the $j^{th}$ prototype feature to represent the information of this query-prototype pair. Specifically, we split both $\mathbf{z}_i$ and $\mathbf{c}_j$ into $L$ groups along their channel dimensions to have $\{z_i^l | l=1,\cdots, L\}$ and $\{c_j^l | l=1,\cdots, L\}$, where $L \leq D$. We calculate the cosine similarity of the $l^{th}$ group as $r_{ij}^l= \frac{z_i^l {c_j^l}^T}{\|z_i^l\|\|c_j^l\|}$, and stack them to be a relation feature vector $\mathbf{v}_{ij}=[r_{ij}^1,\cdots,r_{ij}^L] \in \mathbb{R}^L$. Note that the relations/similarities observed from multiple groups provide valuable hints of the uncertainty since they reflect the similarity from different perspectives.

\begin{table*}[t]
  \centering
    \resizebox{1.0\textwidth}{!}{
    \begin{tabular}{ccccccccccc}
    \toprule
    \multirow{2}[0]{*}{Model} & \multirow{2}[0]{*}{Stage1} & \multirow{2}[0]{*}{Stage2} & \multicolumn{2}{c}{mini-ImageNet} & \multicolumn{2}{c}{tiered-ImageNet} & \multicolumn{2}{c}{CIFAR-FS} & \multicolumn{2}{c}{FC-100} \bigstrut[b]\\
    \cline{4-11}
          &       &       & 1-shot & 5-shot & 1-shot & 5-shot & 1-shot & 5-shot & 1-shot & 5-shot \bigstrut[t]\\
    \hline
    1     & w/o U    & no  & 59.28 $\pm$ 0.81 & 78.26 $\pm$ 0.60 & 67.02 $\pm$ 0.84 & 83.56 $\pm$ 0.61 & 71.95 $\pm$ 0.81 & 84.21 $\pm$ 0.58 & 39.42 $\pm$ 0.77 & 54.12 $\pm$ 0.79 \bigstrut[t]\\
    2     & w U   & no  & 61.53 $\pm$ 0.83 & 78.83 $\pm$ 0.60 & 68.77 $\pm$ 0.82 & 83.92 $\pm$ 0.42 & 73.81 $\pm$ 0.78 & 84.54 $\pm$ 0.63 & 41.76 $\pm$ 0.52 & 56.63 $\pm$ 0.24 \\
    \hline
    3 (Meta-Base)     & w/o U    & w/o U    & 63.10 $\pm$ 0.85 & 79.44 $\pm$ 0.65 & 67.72 $\pm$ 0.80 & 83.61 $\pm$ 0.62 & 72.36 $\pm$ 0.67 & 84.43 $\pm$ 0.50 & 40.23 $\pm$ 0.22 & 56.16 $\pm$ 0.56 \\
    4     & w U   & w/o U    & 63.16 $\pm$ 0.66 & 79.56 $\pm$ 0.67 & 68.92 $\pm$ 0.77 & 83.96 $\pm$ 0.56 & 73.10 $\pm$ 0.70 & 84.68 $\pm$ 0.48 & 40.34 $\pm$ 0.59 & 56.24 $\pm$ 0.55 \\
    5     & w/o U    & w U   & 62.57 $\pm$ 0.33 & 79.35 $\pm$ 0.49 & 69.11 $\pm$ 0.83 & 84.12 $\pm$ 0.55 & 72.51 $\pm$ 0.68 & 84.46 $\pm$ 0.49 & 39.49 $\pm$ 0.33 & 53.24 $\pm$ 0.24 \\
    \hline
    6 (Meta-UAFS)   & w U   & w U   & \textbf{64.22 $\pm$ 0.67} & \textbf{79.99 $\pm$ 0.49} & \textbf{69.13 $\pm$ 0.84} & \textbf{84.33 $\pm$ 0.59} & \textbf{74.08 $\pm$ 0.72} & \textbf{85.92 $\pm$ 0.42} & \textbf{41.99 $\pm$ 0.58} & \textbf{57.43 $\pm$ 0.38} \\
    \bottomrule
    \end{tabular}}%
  \caption{Comparison on applying the proposed similarity uncertainty modeling and optimization in different training stages. ``Stage1'' refers to the \textit{classification-based pre-training} stage while ``Stage2'' refers to the \textit{meta-learning} stage as described in the manuscript. ``no'' denotes the corresponding training stage is not used. ``w U" denotes that we apply the proposed similarity \emph{uncertainty} modeling and optimization (UMO) in the corresponding training stage while the ``w/o U" denotes we do not apply it. }
  \label{tab:training}%
  \vspace{-2mm}
\end{table*}%

The $N$ nodes in the graph are represented by a matrix ${V}_i=(\mathbf{v}_{i1};\cdots;\mathbf{v}_{iN}) \in \mathbb{R}^{N\times L}$. We use graph convolutional network (GCN) to learn the similarity uncertainty for each node. Similar to \cite{shi2019two,wang2018non,wang2018videos}, the edge from the $j^{th}$ node to the $j'^{~th}$ node is modeled by their affinity in the embedded space, which is formulated as:
\begin{equation}\label{eq:6}
E_i(j, j')= \varphi_1(\mathbf{v}_{ij})\varphi_2(\mathbf{v}_{ij'})^T,
\end{equation}
where $\varphi_1$ and $\varphi_2$ denote two embedding functions implemented by FC layers. All edges constitute an adjacent matrix denoted by $E_i \in \mathbb{R}^{N \times N}$. We normalize each row of $E_i$ with SoftMax function so that all the edge values connected to a target node is 1, yielding a numerically-stable message passing through the modeled graph. Similar to \cite{wang2018non}, we denote the normalized adjacent matrix by $G_i$ and update the nodes through GCN as:
\begin{equation}\label{eq:7}
V_i' = V_i + Y_i W_y, ~~where ~~Y_i= G_i V_i W_{\upsilon},
\end{equation}
$W_{\upsilon} \in \mathbb{R}^{L \times L}$ and $W_y \in \mathbb{R}^{L \times L}$ are two learnable transformation matrices. $W_{\upsilon}$ is implemented by a $1 \times 1$ convolutional layer.  $W_y$ is implemented by two stacked blocks. Each block consists of a $1 \times 1$ convolution layer, followed by a Batch Normalization (BN) layer and an LeakyReLU activation layer. We thus infer the similarity uncertainty vector $\mathbf{u}_i = [\sigma_{i1},\cdots,\sigma_{iN}] \in \mathbb{R}^{N}$ for the $N$ pairs from the updated graph nodes, which can be formulated as:
\begin{equation}\label{eq:8}
\mathbf{u}_i = \alpha(\mathrm{BN}(V_i'W_{u1}))W_{u2},
\end{equation}
where the $j^{th}$-dimension of $\mathbf{u}_i$ is the aforementioned $\sigma_{ij}$, denoting the similarity uncertainty for the query $\textbf{x}_{i}$ and the $j^{th}$ prototype. $W_{u1} \in \mathbb{R}^{L \times L}$ and $W_{u2} \in \mathbb{R}^{L \times 1}$ are transformation matrices both implemented by a $1 \times 1$ convolutional layer. ``BN'' refers to the Batch Normalization layer and $\alpha(\cdot)$ refers to the LeakyReLU activation function.

\emph{Discussion:} We model pairwise uncertainty for the similarity instead of for a sample itself like in other tasks \cite{kendall2017uncertainties}. This is because the classification of a query image is influenced by not only the query itself but also the prototypes in the support set. We further conduct extensive experiments to compare the effectiveness of modeling feature uncertainty (query itself) vs. similarity uncertainty (query and prototypes) for few-shot image classification in our experiments. Unlike the previous work \cite{garcia2017few}, which adopts graph-based model to propagate information from labeled samples to unlabeled ones thus delivers a transductive solution, we use graph-based model to achieve the joint estimation of uncertainty of $N$ query-prototype pairs for a given query (non-transductive), where the uncertainty is used to better optimize the network. We do not introduce any complexity increase in the testing where the GCN-based uncertainty estimator is discarded. In contrast, the GCN in \cite{garcia2017few} is a part of their network and is needed during testing.

\paragraph{Joint Training.} Similar to the baseline configuration described in Section 3.1, we train the entire network in two stages: \textit{classification-based pre-training} and \textit{meta-learning}.

In the \textit{classification-based pre-training} stage, the FC coefficients can be considered as the prototypes and the number of input nodes in GCN is the total number of classes in $\mathcal{C}_{base}$. In the \textit{meta-learning} stage, the number of input nodes in GCN is $N$. Because our graph-based uncertainty estimator has the scalability to the number of involved categories for classification, the parameters in GCN are shared in the two stages. We fine-tune the backbone and the similarity uncertainty estimator in the \textit{meta training} stage.

\section{Experiments}

\subsection{Dataset and Implementation}

\paragraph{Datasets.} For few-shot image classification, we conduct experiments on four public benchmark datasets: mini-ImageNet~\cite{vinyals2016matching}, tiered-ImageNet~\cite{ren2018meta}, CIFAR-FS~\cite{bertinetto2018meta}, and FC100~\cite{oreshkin2018tadam}.

\paragraph{Networks and Training.} Following recent common practices \cite{oreshkin2018tadam,lee2019meta}, we adopt a wider ResNet-12 with more channels as the backbone in our work unless claimed otherwise. We build our strong baseline (see Section 3.1) with two-stage training by following \cite{chen2020new}. For all our models, the classification temperature $\tau$ in (\ref{eq:2}) is a trainable parameter which is initialized with 10. For the parameter $L$, we found the performance is very similar when it is in the range of 16 to 64 and we set it to 32.

\paragraph{Evaluation.} We conduct all experiments in the inductive (non-transductive) setting (\ieno, each query sample is evaluated independently from other query samples). We create each few-shot episode by uniformly sampling 5 classes (\ieno $N$=5) from the test set and further uniformly sampling support and query samples for each sampled class accordingly. Consistent with other works, we report the mean accuracy and the standard deviation with a 95\% confidence interval over 1000 randomly sampled few-shot episodes. 

\begin{table}[t]
  \centering
    \resizebox{0.48\textwidth}{!}{
    \begin{tabular}{rrccrr}
    \toprule
    \multicolumn{1}{c}{\multirow{2}[0]{*}{Methods}} & \multicolumn{1}{c}{\multirow{2}[0]{*}{U-Estimator}} & \multicolumn{2}{c}{mini-ImageNet} & \multicolumn{2}{c}{CIFAR-FS} \bigstrut\\
    \cline{3-6}
    \multicolumn{1}{c}{} & \multicolumn{1}{c}{} & 1-shot & 5-shot & \multicolumn{1}{c}{1-shot} & \multicolumn{1}{c}{5-shot} \bigstrut[t]\\
    \hline
    \multicolumn{1}{l}{Meta-Base} & \multicolumn{1}{c}{-} & 63.10 $\pm$ 0.85 & 79.44 $\pm$ 0.65 & \multicolumn{1}{c}{72.36 $\pm$ 0.67} & \multicolumn{1}{c}{84.43 $\pm$ 0.50} \bigstrut[t]\\
    \hline
    \multicolumn{1}{l}{B+ SampU} & \multicolumn{1}{c}{Conv-based} & 63.25 $\pm$ 0.81 & 79.39 $\pm$ 0.63 &
    \multicolumn{1}{c}{72.11 $\pm$ 0.80} &  84.26 $\pm$ 0.52 \\
    \hline
    \multicolumn{1}{l}{B+ SimiU} & \multicolumn{1}{c}{FC-based} & 57.42 $\pm$ 0.80 & 77.01 $\pm$ 0.66 & \multicolumn{1}{c}{70.33 $\pm$ 0.84} & \multicolumn{1}{c}{79.58 $\pm$ 0.78} \bigstrut[t]\\
    \multicolumn{1}{l}{B+ SimiU} & \multicolumn{1}{c}{Conv-based} & 63.31 $\pm$ 0.68 & 79.63 $\pm$ 0.46 & \multicolumn{1}{c}{73.22 $\pm$ 0.75} & \multicolumn{1}{c}{84.40 $\pm$ 0.69} \\
    \multicolumn{1}{l}{B+ SimiU} & \multicolumn{1}{c}{Graph-based} & \textbf{64.22 $\pm$ 0.67} & \textbf{79.99 $\pm$ 0.49} & \multicolumn{1}{c}{\textbf{74.08 $\pm$ 0.72}} & \multicolumn{1}{c}{\textbf{85.92 $\pm$ 0.42}} \\
    \bottomrule
    \end{tabular}}%
  \caption{Comparison on different uncertainty modelling methods. ``SampU'' denotes that we estimate the uncertainty of \emph{a query sample} while ``SimiU'' denotes that we estimate the uncertainty of \emph{the similarity between a query and its prototype}. Besides, we compare different designs of the uncertainty estimator (U-Estimator), including using CNNs (``Conv-based'') to process each sample or each query-prototype pair independently, using GCNs (``Graph-based'') to process the $N$ query-prototype pairs jointly. ``B" denotes the abbreviation of ``Meta-Base".}
  \label{tab:analysis}%
  \vspace{-3mm}
\end{table}%

\begin{table*}[t!]
  \centering
    \vspace{-1mm}
    \resizebox{1.0\textwidth}{!}{
    \begin{tabular}{rccccccccc}
    \toprule
    \multicolumn{1}{c}{\multirow{2}[0]{*}{Methods}} & \multirow{2}[0]{*}{Backbone} & \multicolumn{2}{c}{mini-ImageNet} & \multicolumn{2}{c}{tiered-ImageNet} & \multicolumn{2}{c}{CIFAR-FS} & \multicolumn{2}{c}{FC-100} \bigstrut[b]\\
    \cline{3-10}
    \multicolumn{1}{c}{} &       & 1-shot & 5-shot & 1-shot & 5-shot & 1-shot & 5-shot & 1-shot & 5-shot \bigstrut[t]\\
    \hline
    \multicolumn{1}{l}{MatchingNet \cite{vinyals2016matching}} & 64-64-64-64 & 46.6  & 60.0   &   -   &   -   &   -   &   -   &   -   &  - \bigstrut[t]\\
    \multicolumn{1}{l}{MAML \cite{finn2017model}} & 32-32-32-32 & 48.70 $\pm$ 1.84 & 63.11 $\pm$ 0.92 & 51.67 $\pm$ 1.81 & 70.30 $\pm$ 1.75 &   -   &   -   &  -     &  - \bigstrut[t]\\
    \multicolumn{1}{l}{ProtoNet$^\dagger$ \cite{snell2017prototypical}} & 64-64-64-64 & 49.42 $\pm$ 0.78 & 68.20 $\pm$ 0.66 & 53.31 $\pm$ 0.89 & 72.69 $\pm$ 0.74 &  -      &   -   &   -   &  -  \\
    \multicolumn{1}{l}{RelationNet \cite{sung2018learning}} & 64-96-128-256 & 50.44 $\pm$ 0.82 & 65.32 $\pm$ 0.70 & 54.48 $\pm$ 0.93 & 71.32 $\pm$ 0.78 &   -   & -      &   -   &   -  \\
    \multicolumn{1}{l}{TADAM \cite{oreshkin2018tadam}} & ResNet-12 & 58.50 $\pm$ 0.30 & 76.70 $\pm$ 0.30 &   -   &    -  &   -   &   -   & 40.10 $\pm$ 0.40  & 56.10 $\pm$ 0.40 \\
    \multicolumn{1}{l}{LEO$^\dagger$ \cite{rusu2018meta}} & WRN-28-10 & 61.76 $\pm$ 0.08 & 77.59 $\pm$ 0.10 &  66.33 $\pm$ 0.05  & 81.44 $\pm$ 0.09 &   -   &    -  &   -   &  - \\
    \multicolumn{1}{l}{TapNet \cite{yoon2019tapnet}} & ResNet-12 & 61.65 $\pm$ 0.15 & 76.36 $\pm$ 0.10 & 63.08 $\pm$ 0.15  & 80.26 $\pm$ 0.12 &   -   &   -   &   -   &   -  \\
    \multicolumn{1}{l}{Shot-free \cite{ravichandran2019few}} & ResNet-12 & 59.04 $\pm$ 0.43 & 77.64 $\pm$ 0:39 & 66.87 $\pm$ 0.43 & 82.64 $\pm$ 0.39 & 69.20 $\pm$ 0.40 & \underline{84.70 $\pm$ 0.40} &   -    &  -  \\
    \multicolumn{1}{l}{MetaOptNet \cite{lee2019meta}} & ResNet-12 & 62.64 $\pm$ 0.61 & 78.63 $\pm$ 0.46 & 65.99 $\pm$ 0.72 & 81.56 $\pm$ 0.53 & 72.00 $\pm$ 0.70 & 84.20 $\pm$ 0.50 & \underline{41.10 $\pm$ 0.60} & 55.50 $\pm$ 0.60 \\
    \multicolumn{1}{l}{CAN \cite{hou2019cross}} & ResNet-12 & \underline{63.85 $\pm$ 0.48} & \underline{79.44 $\pm$ 0.34} & \textbf{69.89 $\pm$ 0.51} & \underline{84.23 $\pm$ 0.37} & - & - & - & - \\
    \multicolumn{1}{l}{Baseline2020 \cite{dhillon2019baseline}} & WRN-28-10 & 56.17 $\pm$ 0.64 & 73.31 $\pm$ 0.53 & 67.45 $\pm$ 0.70 &  82.88 $\pm$ 0.53  & 70.26 $\pm$ 0.70  & 83.82 $\pm$ 0.49 & 36.82 $\pm$ 0.51 & 49.72 $\pm$ 0.55 \\
    \multicolumn{1}{l}{MetaBaseline \cite{chen2020new}$^1$ } & ResNet-12 & 63.17 $\pm$ 0.23 & 79.26 $\pm$ 0.17 & 68.62 $\pm$ 0.27 & 83.29 $\pm$ 0.18 &   -   &  -     &   -    &  - \\
    \hline
    \multicolumn{1}{l}{Meta-Base} & ResNet-12 & 63.10 $\pm$ 0.85 & 79.44 $\pm$ 0.65 & 67.72 $\pm$ 0.80 & 83.61 $\pm$ 0.62 & \underline{72.36 $\pm$ 0.67} & 84.43 $\pm$ 0.50 & 40.23 $\pm$ 0.22 & \underline{56.16 $\pm$ 0.56} \bigstrut[t]\\
    \multicolumn{1}{l}{Meta-UAFS} & ResNet-12 & \textbf{64.22 $\pm$ 0.67} & \textbf{79.99 $\pm$ 0.49} & \underline{69.13 $\pm$ 0.84} & \textbf{84.33 $\pm$ 0.59} & \textbf{74.08 $\pm$ 0.72} & \textbf{85.92 $\pm$ 0.42} & \textbf{41.99 $\pm$ 0.58} & \textbf{57.43 $\pm$ 0.38} \\
    \bottomrule
    \end{tabular}}%
  \caption{Accuracy (\%) comparison for 5-way few-shot classification of our schemes, our baselines and the state-of-the-arts on four benchmark datasets. For the backbone network, ``$l_1$-$l_2$-$l_3$-$l_4$'' denotes a 4-layer convolurional network with the number of convolutional filters in each layer, respectively. ``Meta-Base'' and ``Meta-UAFS'' refer to the strong baseline model and our final UAFS model trained with \textit{classification-based pre-training} stage and \textit{meta-learning} stage, respectively. The superscript $^\dagger$ refers to that the model is trained on the combination of training and validation sets while others only use the training set. Note that for fair comparisons, all the presented results are from inductive (non-transductive) setting. Bold numbers denotes the best performance while numbers with underlines denotes the second best performance.}
  \label{tab:SOTA&Ablation}%
  \vspace{-3mm}
\end{table*}%

\subsection{Ablation Study}
\paragraph{Effectiveness of Proposed UAFS.} Table \ref{tab:training} shows the ablation study on the effectiveness of our proposed method. Model-3 corresponds to our strong baseline \emph{Meta-Base} (see Table \ref{tab:SOTA&Ablation}) with two-stage training. Model-6 corresponds to our final scheme \emph{Meta-UAFS} with the proposed similarity uncertainty modeling and optimization (UMO).

We have the following observations. \textbf{1)} Our proposed \emph{Meta-UAFS} (Model-6) improves the strong baseline \emph{Meta-Base} (Model-3) by 1.12\%, 1.41\%, 1.72\% and 1.76\% in 1-shot classification on four datasets respectively, which demonstrates the effectiveness of our UMO. \textbf{2)} \emph{Meta-Base} (Model-3) obviously outperforms Model-1 that  has only Stage1 training, demonstrating the effectiveness of two-stage training. \textbf{3)} Enabling UMO in the \emph{classification-based pre-training stage} (Stage1) brings obvious improvement for one-stage training (Model-2 vs. Model-1), but only slight gains for two-stage training (Model-4 vs. Model-3). The higher the performance, the more difficult it is to obtain gain. On top of the strong baseline, the UMO is not helpful for Stage1 which does not suffer from limited training data as Stage2. \textbf{4)} Adopting UMO only in Stage2 (Model-5) does not always bring improvements (when compared with \emph{Meta-Base}), while applying UMO on both stages (\emph{Meta-UAFS}) brings significant improvements. For Model-5, the uncertainty estimator is trained only in Stage2 (from scratch without pre-training in Stage1) and may be difficult to optimize. Our graph-based uncertainty estimator has the scalability to the number of involved categories for classification, which allows us to pre-train the parameter-sharable uncertainty estimator in Stage1 and fine-tune it in Stage2. When we enable the pre-training of uncertainty estimator in Stage1, our final scheme (\emph{Meta-UAFS}) consistently delivers the best results. Note that the backbone network is trained in both stages.

\paragraph{Similarity Uncertainty Estimator Designs.}
Since the query is classified in a joint determination by comparing it with the $N$ prototypes simultaneously, we propose to use a GCN to jointly estimate the uncertainties. This design is superior because it 1) enables the joint estimation of similarity uncertainties and 2) has the scalability to the number of categories which facilitates the pre-training of the uncertainty estimator in Stage-1. Table \ref{tab:analysis} shows that 1) our \emph{Graph-based} design (last row) is superior to \emph{Conv-based} design (the penultimate row) which uses $1\times1$ convolutional layers to independently infer the uncertainty for each query-prototype pair. FC-based design is inferior to our proposed design.

\paragraph{Sample Uncertainty vs. Pair Similarity Uncertainty.} Table \ref{tab:analysis} shows that the scheme \emph{B+SimiU} with Graph-based uncertainty estimator which models the \emph{uncertainty of the query-prototype similarity} is more effective than the scheme \emph{B+SampU} which models the  \emph{uncertainty of a sample}. That is because in the few shot classification, for a given query, its estimated class depends on both the query and the $N$ prototypes. Therefore, we model the uncertainty of the similarity between the query and the prototypes.

\subsection{Comparison with State-of-the-arts}

In Table  \ref{tab:SOTA&Ablation}, we compare our proposed  \emph{Meta-UAFS} with the state-of-the-art approaches. Note that our baseline \emph{Meta-Base} is the same as MetaBaseline \cite{chen2020new} and they have the similar performance. Compared to \emph{Meta-Base}~\cite{chen2020new}, our \emph{Meta-UAFS} achieves significant improvement of 1.12\%, 1.41\%, 1.72\%, and 1.76\% in 1-shot accuracy on mini-ImageNet, tiered-ImageNet, CIFAR-FS, and FC00, respectively for 1-shot classification. Our \emph{Meta-UAFS} achieves the state-of-the-art performance. CAN \cite{hou2019cross} introduces cross attention module to highlight the target object regions, making the extracted feature more discriminative. Our uncertainty modeling and optimization which reduces the side effects of observation noises is complementary to their attention design and incorporating their attention design into our framework would lead to superior performance.

\vspace{-1mm}
\section{Conclusion}

In this paper, we propose an Uncertainty-Aware Few-Shot image classification approach where data-dependent uncertainty modeling is introduced to alleviate the side effect of observation noise. Particularly, we convert the similarity between a query sample and a prototype from a deterministic value to a probabilistic representation and optimize the networks by exploiting the similarity uncertainty. We design a graph-based uncertainty estimator to jointly estimate the similarity uncertainty of the query-support pairs for a given query sample. Extensive experiments demonstrate the effectiveness of our proposed method and our scheme achieves state-of-the-arts performance on all the four benchmark datasets.

\section*{Acknowledgements} This work was supported in part by the National Key Research and Development Program of China 2018AAA01014-\\00 and NSFC under Grant U1908209, 61632001.

\small
\bibliographystyle{named}
\bibliography{ijcai21}

\end{document}